\documentclass[letterpaper, 10 pt, conference]{ieeeconf}  

\IEEEoverridecommandlockouts                              

\usepackage{graphics} 
\usepackage{epsfig} 
\usepackage{amsmath} 
\usepackage{siunitx}
\usepackage{graphicx}
\usepackage{subfigure}
\usepackage{tikz}

\title{\LARGE \bf
Soft Adaptive Feet for Legged Robots: An Open-Source Model for Locomotion Simulation
}

\author{Matteo Crotti$^{1,2}$, Luca Rossini$^{3}$, Balint K. Hodossy$^{4}$, Anna Pace$^{1}$, \\ Giorgio Grioli$^{1,2}$, Antonio Bicchi$^{1,2}$,  Manuel G. Catalano$^{1}$
\thanks{This research has received funding from the European Union’s Horizon 2020 Research, ERC programme under the Grant Agreement No.810346 (Natural BionicS).}
\thanks{$^{(1)}$ Soft Robotics for Human Cooperation and Rehabilitation, Fondazione Istituto Italiano di Tecnologia, Genoa 16163, Italy.}
\thanks{$^{(2)}$ Department of Information Engineering and Research Center “E. Piaggio,” University of Pisa, Pisa 56122, Italy.}%
\thanks{$^{(3)}$ Humanoids and Human Centered Mechatronics, Fondazione Istituto Italiano di Tecnologia, Genoa 16163, Italy.}%
\thanks{$^{(4)}$ Department of Bioengineering, Imperial College London, SW7 2AZ London, U.K.}%
\thanks{This letter has supplementary downloadable material available at https://ieeexplore.ieee.org, provided by the authors.}
}

\newcommand\copyrighttext{%
  \footnotesize \textcopyright\ \the\year{} This work has been submitted to the IEEE for possible publication. Copyright may be transferred without notice, after which this version may no longer be accessible.}

\newcommand\copyrightnotice{%
\begin{tikzpicture}[remember picture,overlay]
\node[anchor=south,yshift=10pt] at (current page.south) {\fbox{\parbox{\dimexpr\textwidth-\fboxsep-\fboxrule\relax}{\copyrighttext}}};
\end{tikzpicture}%
}

\begin{document}

\maketitle
\thispagestyle{empty}
\pagestyle{empty}

\begin{abstract}

In recent years, artificial feet based on soft robotics and under-actuation principles emerged to improve mobility on challenging terrains. This paper presents the application of the MuJoCo physics engine to realize a digital twin of an adaptive soft foot developed for use with legged robots. 
We release the MuJoCo soft foot digital twin as open source to allow users and researchers to explore new approaches to locomotion.
The work includes the system modeling techniques along with the kinematic and dynamic attributes involved. Validation is conducted through a rigorous comparison with bench tests on a physical prototype, replicating these experiments in simulation. Results are evaluated based on sole deformation and contact forces during foot-obstacle interaction. 
The foot model is subsequently integrated into simulations of the humanoid robot COMAN+, replacing its original flat feet. Results show an improvement in the robot's ability to negotiate small obstacles without altering its control strategy.
Ultimately, this study offers a comprehensive modeling approach for adaptive soft feet, supported by qualitative comparisons of bipedal locomotion with state of the art robotic feet.

\end{abstract}

\copyrightnotice
\section{INTRODUCTION}

Replicating human locomotion and stability on uneven terrains has been a longstanding challenge in robotics. Flat-footed (e.g., ATLAS, Digit, HRP-4 \cite{Caron2019}, G1, Nao, Talos), and passive ball-shaped (e.g., ANYmal \cite{Hutter2016}, HyQ \cite{Boaventura2013}, Spot) robots often struggle to traverse irregular terrains, limiting their applications in unstructured environments. 

Humanoid robotic feet come in various structural designs, generally consisting of either a single rigid segment or two segments linked by a revolute joint. While effective on flat surfaces, these designs often lack the adaptability needed for uneven terrains. To address this limitation, more advanced designs have been introduced to enhance stability and agility. For instance, foot soles incorporating compliant visco-elastic materials \cite{Owaki2012, dema2017sole, Nikos2015} or airbag units \cite{Najmuddin2012, Zang2017} can absorb small ground irregularities. Other designs use articulated structures made up of multiple segments joined by joints and compliant elements \cite{Seo2009ModelingAA, Davis2010TheDO, Inaba2016}. Our team adopted a different approach with the SoftFoot, an adaptive passive robotic foot. This design focuses on ground adaptability at a mechanical level, featuring a deformable sole that replicates the structure of the human foot \cite{Piazza2024}. Testing on the humanoid robot HRP-4 as it stepped on small obstacles showed significant improvements in stability and obstacle negotiation \cite{Catalano2021hrp4}.

\begin{figure}
    \centering
    \includegraphics[width=\columnwidth]{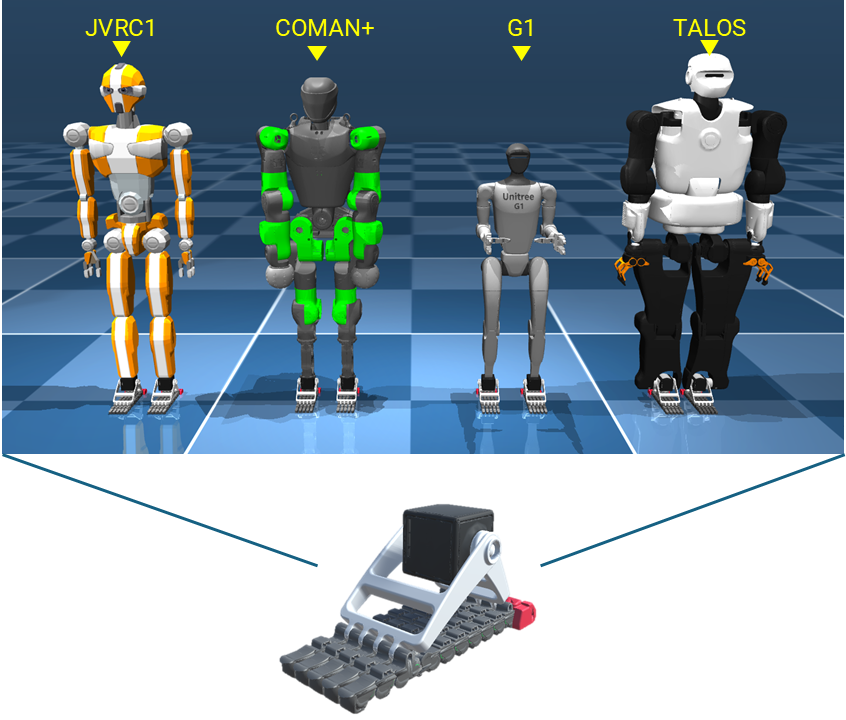} 
    \caption{The SoftFoot digital twin (on the bottom) and its application to various legged robots in MuJoCo. From left to right: JVRC1, COMAN+, G1 from Unitree, and Talos from Pal Robotics.}
    \label{fig:model}
    \vspace{-1em}
\end{figure}


Soft technologies offer enhanced adaptability, but their complexity poses substantial mathematical challenges. Analytical models often require simplifications to handle the dynamic interactions, uncertainties, and nonlinearities inherent in these systems. For example, ball-shaped feet are often reduced to a single ground contact point \cite{Kim2016}; similarly, the SoftFoot analytical model proposed in \cite{Mura2020}, which evaluates contact force distribution, works under specific assumptions.


Physics engines and digital twins offer powerful tools for simulating and testing complex robotic systems \cite{Tammi2019, Li2021}. Physics engines provide dynamic virtual environments that simplify the design process and allow real-time testing and optimization of mechanical components and control strategies \cite{OpenAI2019, Yoon2023, Gentiane2022}.
MuJoCo, known for its speed and accuracy in constrained systems, is especially effective for simulating humanoid robots and their behavior on uneven terrains \cite{Erez2015, Ivaldi2015}.
Digital twins, as virtual replicas of physical systems, support real-time simulation, performance prediction, and iterative refinement. For the SoftFoot, a digital twin enables accurate simulation on uneven terrains, to test the increased adaptability and stability it can offer without extensive physical prototyping.


In this work, we present a digital twin of the robotic SoftFoot developed within the MuJoCo physics engine (Fig. \ref{fig:model}) and its use in locomotion with legged robots. This model allows to analyze the interaction of the SoftFoot with uneven terrains and obstacles, and it can be easily fitted on MuJoCo legged robot models (see Fig. \ref{fig:model}) to investigate the potential of soft adaptive feet in locomotion. We thereby release the SoftFoot model as open-source within the Natural Machine Motion Initiative \cite{dellasantina2017} to allow other users to explore new locomotion approaches.

The paper outlines the model description within the simulator (Section II) and its validation (Section III), conducted through comparison with bench tests on the foot physical prototype replicated in simulation \cite{Mura2020}. The metrics used to assess the model’s accuracy and performance are the foot sole deformation and the contact forces. We then simulate the locomotion of a humanoid robot walking with the SoftFoot and its original feet (section IV), as described in \cite{Catalano2021hrp4}, to qualitatively analyze the capability of the adaptive foot in negotiating obstacles. Finally, we discuss the model accuracy and its effectiveness, underlying its limitations (section V).

\section{SOFTFOOT MODEL}
\subsection{The SoftFoot}

\begin{figure}
    \centering
    \includegraphics[width=\columnwidth]{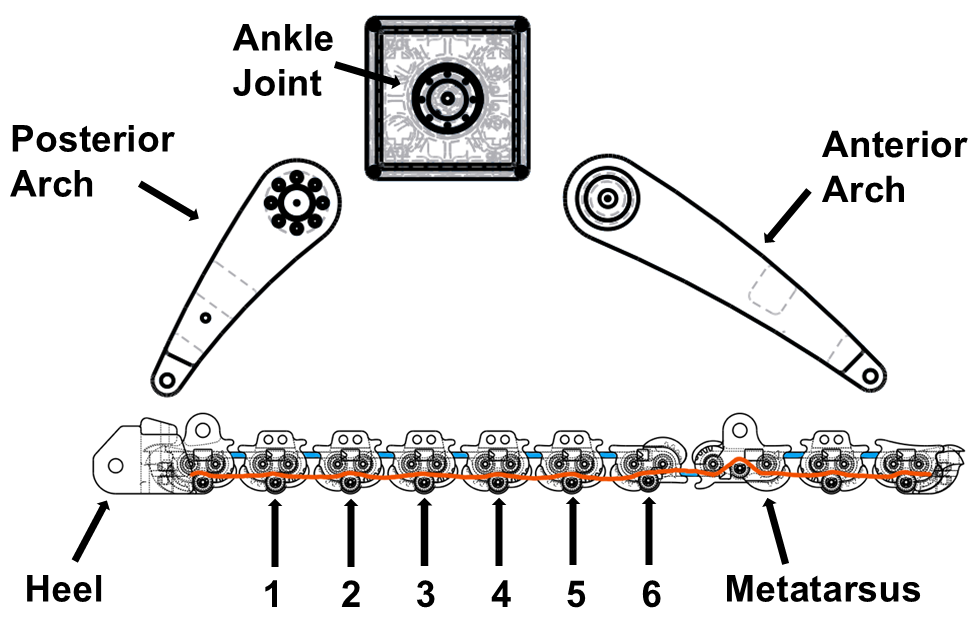}
    \caption{Exploded view of the robotic SoftFoot in the sagittal plane. The order of the plantar fascia modules is reported. The elastic bands in the sole are in blue and the tendon is in red. 
    \label{fig:softfoot_cad}}
\end{figure}

\par 
The SoftFoot is composed of simple, compact, and modular mechanical structures \cite{Piazza2024}.
The SoftFoot features a posterior arch rigidly connected to the ankle joint, and an anterior arch connected to the posterior with a revolute joint (Fig. \ref{fig:softfoot_cad}).
Five modular and flexible parallel structures called \textit{chains}, form the deformable sole, which acts as the \textit{plantar fascia} (a thick band of tissue running across the bottom of the foot from the heel bone to the toes) and \textit{toes} of the human foot. Each of these structures is designed with a series of modular components, which will be referred to as \textit{module} from here onwards, similar to those developed for the Pisa/IIT SoftHand \cite{dellasantina2017}, engineered to roll on each other utilizing a specific flexible configuration of the Rolamite joint. The modules are secured together with two elastic bands for each coupling. A tendon runs through each of the five chains, following a specific route as depicted in Fig. \ref{fig:softfoot_cad}, to distribute forces between joints.

\par The \textit{metatarsus} module is specifically designed to allow the connection of the plantar fascia with the anterior arch through a revolute joint. The device is engineered to replicate the windlass mechanism of the human foot, in which the tightening of the plantar fascia due to toe dorsiflexion elevates the foot arch, providing a stiff lever for propulsion. This mechanism plays a crucial role in the foot's ability to support and move the body forward effectively, particularly during activities such as walking, running, and jumping.

\par The SoftFoot sole changes its shape and stiffness through its system of pulleys, tendons, and springs depending on the applied load and the ground profile. The SoftFoot exhibits compliance when the load is low, stiffening up as the load increases. This architecture is responsible for the foot's adaptiveness, impact absorption, and stabilization capability.

\subsection{The MuJoCo SoftFoot Model}

MuJoCo allows us to model all the SoftFoot components effectively.
First, we gathered all the necessary information from the CAD model of the foot prototype, including the size, relative position, and inertia matrix of each component. We also exported simplified meshes.
Then, we defined the tree structure of the foot model as two open chains (see Fig. \ref{fig:softfoot_mjc}(b)). The first starts from the ankle joint and includes the posterior arch, the heel, and the five chains of the sole up to the tip toes; the second one comprises only the anterior arch.
The two chains are connected, imposing an equality constraint on the position of the metatarsal joint, where the anterior arch and the metatarsus module of each chain connects (see red circle in Fig.\ref{fig:softfoot_mjc}(b)), to form the closed-chain structure between the arches and the foot sole. No constraints are applied to the relative orientation of the two open-chain structures to keep the revolute joint functioning correctly. Finally, we modeled the foot sole components and their functioning:

\begin{figure}[]
    \centering
    \includegraphics[width=.95\columnwidth]{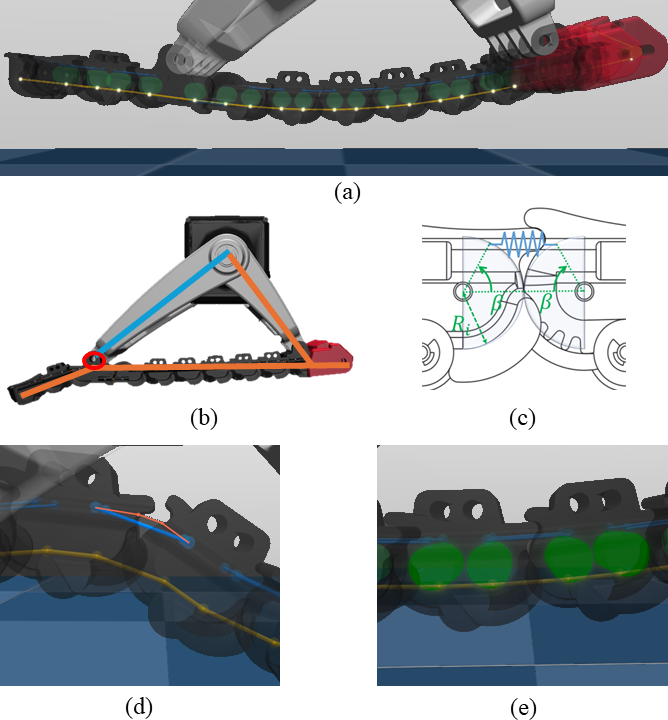}
    \caption{Mujoco techniques to model the SoftFoot. (a) The tendon in the foot sole (in yellow) passes through specific points, a.k.a. sites (in white), defined for each module to reproduce its routing. (b) The two open-chain structures defining the foot model are represented in blue and orange. The red circle illustrates the connection point. (c) Graphical representation of the rotation equality constraint between coupled modules. (d) The elastic band path in the simulation (in blue) goes through the module and slightly differs from the real path (in orange). (e) Close up view of the virtual geometries added to each module as pulleys.}
    \label{fig:softfoot_mjc}
    \vspace{-0em}
\end{figure}

\subsubsection{Modules}
The SoftFoot modules present mechanical limit switches to limit the rotation between couplings. This effect is modeled by imposing a limit in the revolute joint rotation range. Specifically, it has been identified as a rotation range between 20 and 90 degrees, respectively, for the anticlockwise and clockwise rotations. 
The module’s coupling with the Rolamite joint is modeled with the assumption of pure rotation without sliding or losing contact. This is obtained by imposing an equality constraint on the relative rotation between adjacent modules in a coupling (see Fig. \ref{fig:softfoot_mjc}(e)). This assumption also holds for the actual prototype, except under conditions of very high loads, where the modules might lose contact.
This behavior can be replicated by exploiting MuJoCo soft constraints and properly adjusting the parameters of the equality constraints.

\subsubsection{Elastic Bands}
Two distinct approaches are possible to model the elastic bands connecting coupled modules: leveraging MuJoCo's capability to define elastic elements or expressing the bands' properties through joint properties. 
MuJoCo's elastic tendon elements exhibit versatility in replicating the behavior of elastic bands. We configure these elements as unidirectional tension springs with the extremities of the elements fixed within the corresponding coupled modules. As the coupling rotates, the elastic element extends and applies a force to bring the modules back to their resting position. The resting length and lengthening range are derived from datasheets of the bands, while their stiffness value is estimated considering their geometric properties and the elastic modulus estimated with the Gent equation \cite{Gent1958OnTR} starting from the shore A hardness. Damping is heuristically determined.
Alternatively, the elastic band characteristics can be transferred to the coupling joint parameters without modeling additional elastic elements. The joint stiffness is obtained by transposing elastic band stiffness to the correspondent joint angular stiffness.

However, adopting the first modeling approach introduces a gap from the real prototype. Lacking contact with the module itself, the linear elastic element path diverges from the actual behavior of the elastic bands, which follow specific paths defined within the module structure, affecting their stretch as shown in Fig. \ref{fig:softfoot_mjc}(d).
Furthermore, modeling all the elastic elements implies the addition of 45 elements and their 90 extremities in the model, increasing its complexity and computational cost. Therefore, we adopted the second approach.

\subsubsection{Tendons}
We added virtual massless cylindrical geometries in the modules to act as pulleys to accurately replicate the tendon routing and its force distribution functionality (see Fig. \ref{fig:softfoot_mjc}(e)).
The tendon in MuJoCo is defined by the minimal path between designated points.
We set these points in the reference frame of the pulleys so that they rotate accordingly to the module, consequently modifying the contact point between the tendon and the pulley. 
Therefore, the path and the lengths of the tendons are determined by the shape of the ground, i.e. the rotation of the sole modules. 
The prototype’s tendons are almost inextensible, and this behavior is modeled by imposing the admissible tendon length range in the model. The resting length of the tendon element was heuristically obtained from the simulation, and the maximum length was determined considering the lengthening characteristics of the real tendon. Fig. \ref{fig:softfoot_mjc}(a) illustrates one of the tendons in the model and its route along the sole.

MuJoCo uses a soft contact model, allowing the simulated bodies to interpenetrate and generate forces proportional to the depth of contact. 
In our model, the modules bodies are generally in contact due to the action of the elastic bands, creating numerous interpenetration forces, destabilizing the simulation. We therefore excluded inter-module contacts, as these interactions do not impact the foot's functional behavior. The resultant model is shown in the video attachment.

\section{MODEL VALIDATION}
To verify the accuracy of the SoftFoot model, we reproduced in MuJoCo the experiments performed in \cite{Mura2020} to validate an analytical model of the same SoftFoot prototype.


\subsection{Experimental Setup}

In the experiments, a vertical load was applied to a SoftFoot with a single chain sole, which made contact with three sensorized obstacles. Two obstacles were placed under the heel and the metatarsus module, respectively, while a third obstacle, adjustable in height, was positioned to match the six modules of the plantar fascia. This third obstacle was moved to align with the center of each sole module across different trials. Each of the six modules in the plantar fascia (see Fig. \ref{fig:softfoot_cad}) was tested with obstacle heights of \SI{7}{\milli\meter}, \SI{11}{\milli\meter}, \SI{15}{\milli\meter}, and \SI{19}{\milli\meter}. Contact forces at each obstacle were measured using embedded six-axis force/torque sensors, and the rotation of foot elements was tracked with IMUs attached directly to each sole module.

\begin{figure}
    \centering
    \includegraphics[width=0.95\columnwidth]{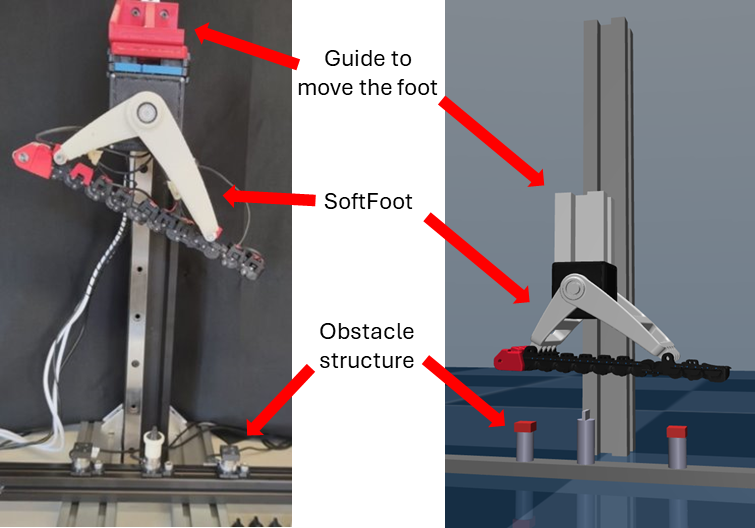}
    \caption{Experimental setup. On the left the setup from the experiments in \cite{Mura2020}, on the right the simulated setup in MuJoCo.}
    \label{fig:setup}
\end{figure}

This experimental setup was recreated in MuJoCo, as illustrated in Fig. \ref{fig:setup}.
The foot was attached to a vertically actuated slider to simulate the applied load. The slider’s initial height was set so that the foot sole had no initial contact with the obstacle. When the simulation began, the combined effects of the load and gravity caused the foot to gradually press down until it reached equilibrium. In this virtual setup, forces exerted by the foot on the obstacles were measured using MuJoCo sensors embedded in each obstacle, while rotations of each foot sole module were obtained from the simulation data. The virtual setup can be seen in the video attachment.

\begin{figure*}[ht]
    \centering
    \includegraphics[width=.95\textwidth]{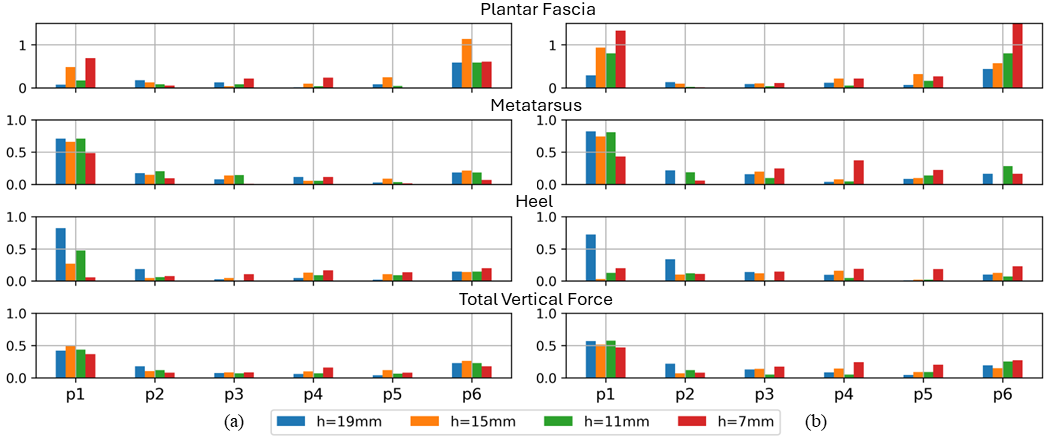}
    \caption{Force relative errors for the three foot parts in contact with the obstacles and the error on the total vertical force. Positions (p) 1 and 6 represent the position closer to the heel and to the metatarsus module respectively. Results are divided according to the applied load, (a) \SI{12}{\newton} and (b) \SI{24}{\newton} respectively.}
    \label{fig:e_rel}
\end{figure*}

\begin{figure*}[ht]
    \centering
    \includegraphics[width=.95\textwidth]{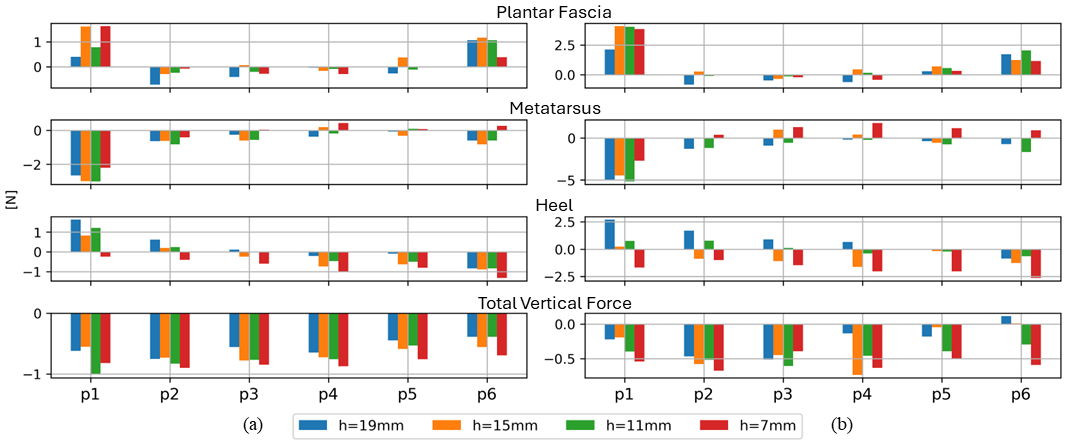}
    \caption{Force errors for the three foot part in contact with the obstacles and the error on the total vertical force. Positions (p) 1 and 6 represent the position closer to the heel and to the metatarsus module respectively. Results are divided according to the applied load, (a) \SI{12}{\newton} and (b) \SI{24}{\newton} respectively.}
    \label{fig:e_abs}
\end{figure*}

\subsection{Methods}
Each trial lasted for a fixed time of 5 seconds, sufficient to reach the equilibrium condition given the initial distance between the SoftFoot and the obstacles. Forces and rotations data from the simulations were retrieved from the last simulation step. 
A total of 48 trials were conducted to replicate all the experimental conditions, which tested two different loads (approximately \SI{12}{\newton} and \SI{24}{\newton}), for four different obstacle heights and six obstacle positions.

The forces obtained from the experimental sessions were compared with those calculated from the simulation. 
Only the vertical component of the force was considered, as it represents the major component of the total force, and the investigation of friction forces in the simulator is out of the scope of this paper.
Relative errors on the forces were computed for each of the three foot components in contact with an obstacle: the heel (\textit{h}), the metatarsus (\textit{m}), and the interested plantar fascia module (\textit{p}). In particular, for each body $b \in \{h, p, m\}$ the relative error on the vertical force was calculated as:

\begin{equation*}
    e_b = (\hat{f_b} - \tilde{f_b})/\tilde{f_b}
\end{equation*}

where $\hat{f_b}$ and $\tilde{f_b}$ are respectively the force measured in the simulation and recorder in the experimental session. 
The error on the total vertical force acting at the three body $b \in \{h, p, m\}$ was calculated as:
\begin{equation*}
    e_T = \sum{(\hat{f_b} - \tilde{f_b})}/\sum{\tilde{f_b}}
\end{equation*}

 
The rotation of the modules in the simulation were compared to those collected during the experiments. We focused on the rotations in the sagittal plane, as the SoftFoot features enhanced ground adaptability in this plane.
The disparity in the initial placement and orientation between the IMUs in the experiments and the reference frames of the bodies in the simulation leads to an offset, which was removed from the simulated rotations to facilitate data comparison. The offset was computed as the mean difference between the simulated and the experimental rotations.

\subsection{Results}

\subsubsection{Force Validation}

Fig. \ref{fig:e_rel} presents the absolute values of the relative errors in the forces measured on the three obstacles under each experimental condition for the two applied loads. Conversely, Fig. \ref{fig:e_abs} illustrates the absolute errors, defined as the difference between the forces in the simulation and those recorded in the experiments. A positive value indicates that the force in the simulation is higher than the experimental, while a negative value implies that it is lower.

\subsubsection{Sole Deformation Validation}

\begin{figure*}
    \centering
    \includegraphics[width=.95\textwidth]{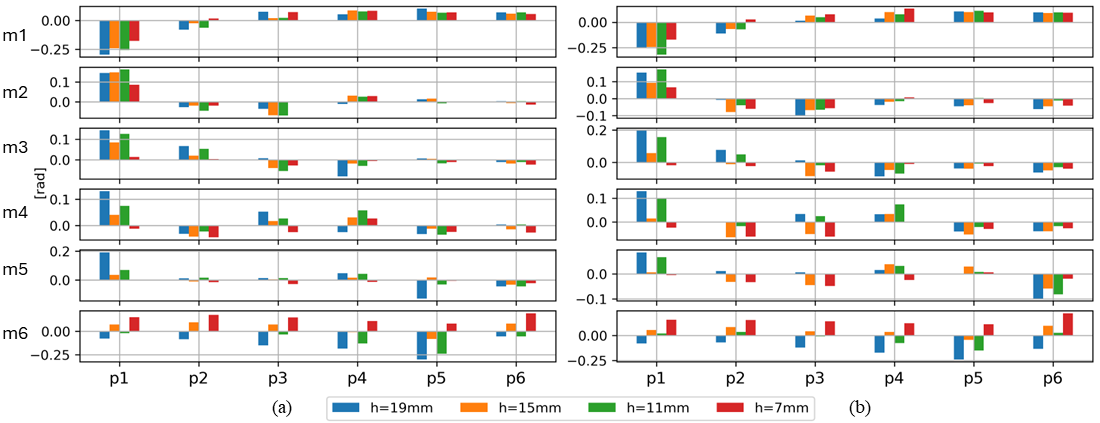}
    \caption{Rotation errors for the different modules composing the plantar fascia (m1-6). Positions (p) 1 and 6 represent the position closer to the heel and to the metatarsus module respectively. Results are divided according to the applied load, (a) \SI{12}{\newton} and (b) \SI{24}{\newton} respectively.}
    \label{fig:e_imu}
\end{figure*}

\begin{figure}
    \centering
    \includegraphics[width=0.46\textwidth]{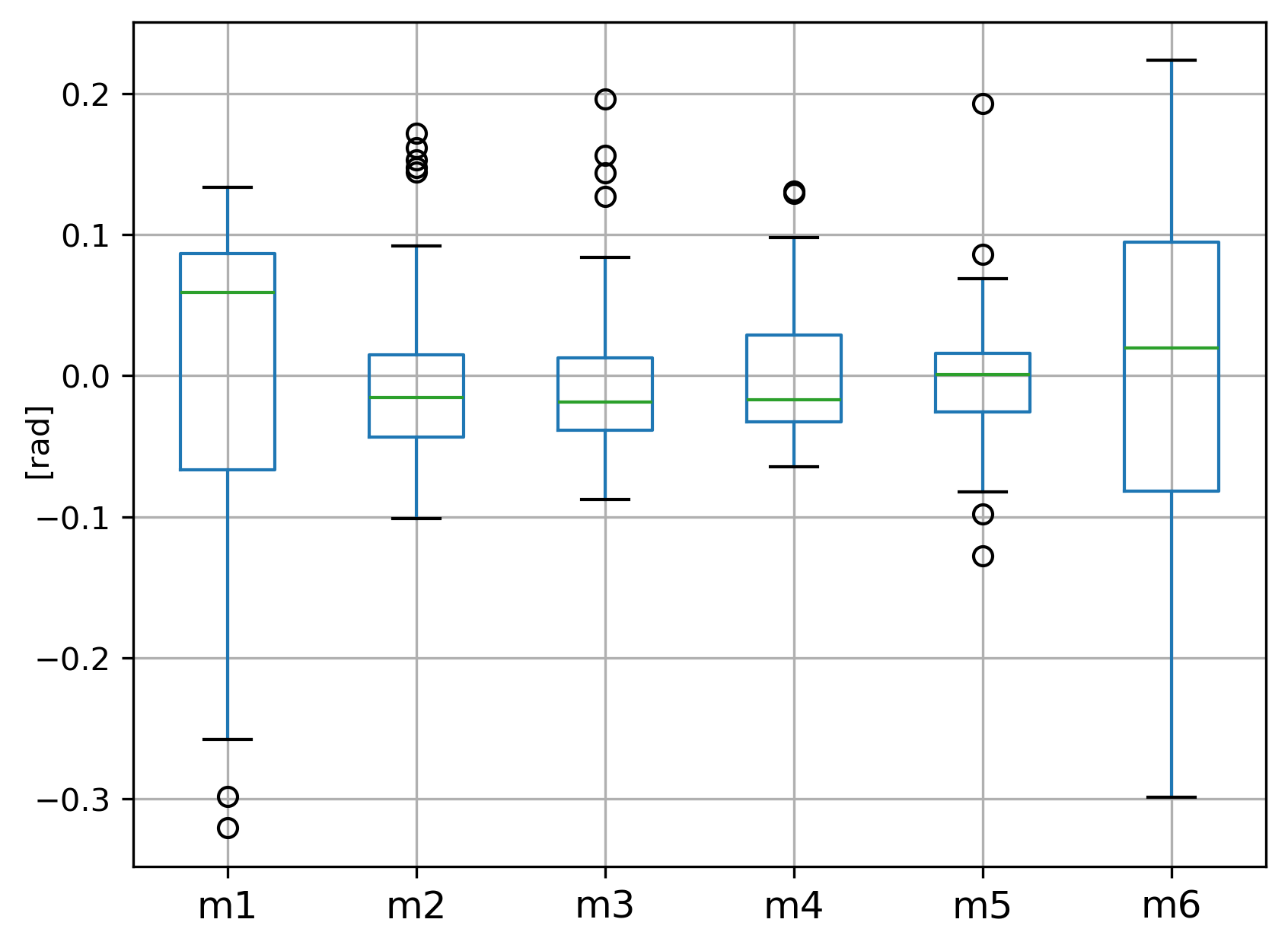}
    \caption{Box plot representation of the rotation error for each of the six plantar fascia modules (m1-6) for all the 48 trials. The green line represents the median, and the box is the interval between the first and the third quartile. The black circles correspond to the outliers.}
    \label{fig:e_imu_box}
\end{figure}

Fig. \ref{fig:e_imu} displays the discrepancies in the rotation of the different plantar fascia modules in the sagittal plane between the experimental data and the simulations, with the results divided according to the applied load. Fig. \ref{fig:e_imu_box}  shows the statistical distribution of rotation errors for each module across all simulations.

\section{BIPEDAL ROBOT LOCOMOTION}
 
To demonstrate the usability and effectiveness of the SoftFoot model, we utilized the torque-controlled humanoid robot COMAN+ \cite{FERRARI2023} to simulate walking over obstacles with squared section, whose side measures \SI{5}{\milli\meter} and \SI{10}{\milli\meter}. These simulations were conducted using both the original flat rigid feet and the SoftFoot, without modifying the robot's control structure. This approach was inspired by the previous work done with the robot HRP-4  \cite{Catalano2021hrp4}. The obstacles were placed in randomized positions for all the tests done.

The robot walks with a whole-body non-linear Model Predictive Control (MPC) using the Horizon framework~\cite{ruscelli:horizon} that generates position, velocities, and torque references, minimizing a cost function that takes into account Cartesian references. 
By setting the Cartesian references and setting the gait sequence and timing, the MPC outputs directly the joint space references that are interpolated to match the low-level control frequency (i.e., 1 kHz).

To simplify ground collisions and improve simulation stability, we added box geometries encapsulating the modules as colliders. Nevertheless, given the high number of contacts with the ground due to the articulated structure of the SoftFoot and the constraints used to model its behavior, it was necessary to decrease the simulation timestep to \SI{0.1}{\milli\second}. 

Differently from the experiments performed for HRP-4, the SoftFoot was not specifically designed for COMAN+, and we just replaced the humanoid original feet with a pair of SoftFoot. The dimensions of our foot are significantly different from those of the robot original foot. The rigid foot is about \SI{21}{\centi\meter} long and \SI{11}{\centi\meter} large, while the SoftFoot has a total length of \SI{27}{\centi\meter}, of which \SI{16}{\centi\meter} are within the two arches, and its width is \SI{9}{\centi\meter}. Moreover, our foot is not symmetrical with respect to the frontal plane.
Given these differences, it was necessary to move the relative position of the robot's ankle with respect to the SoftFoot ankle joint of \SI{1}{\centi\meter} posteriorly, and \SI{2}{\centi\meter} downward to make the robot stand up on the pair of SoftFoot.
We refer the readers to the video attachment for a more detailed view of the simulation.

\subsection{Results}

\begin{figure}[h]
    \centering
    \includegraphics[width=0.95\linewidth]{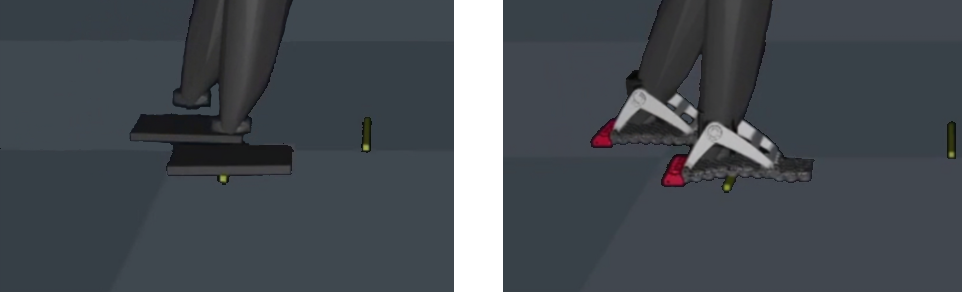}
    \caption{Close-up view on COMAN+ addressing the obstacle with its rigid foot on the left and the SoftFoot on the right.}
    \label{fig:zoom}
\end{figure}

The simulations show that the humanoid robot can overcome small obstacles without requiring a specific control structure when using the SoftFoot, whereas it tends to fall when using its original rigid flat foot. This difference is further depicted in Fig. \ref{fig:zoom}, which shows how the two types of feet interact with the obstacle. While the rigid foot generates a very small contact surface with the obstacle and the ground, resulting in a small support area, the SoftFoot filters the obstacle, optimizing the contact surface and the support area.

\section{DISCUSSION}

\subsection{Force Validation}
The absolute error in the total vertical force remains consistent across different conditions and below 10\% of the applied load. 
The discrepancy in the total vertical force is primarily attributed to a lower total force observed in the simulations. We applied, indeed, the nominal force declared in the experiment, while in the experimental setup, the applied load was affected by the positioning of a physical weight leading to variability in the actual load.

The force error exhibits a consistent pattern across the components of the foot. The error tends to be higher when the moving obstacle is closer to the heel (p1) for all three sole parts. As observed in Fig. \ref{fig:e_abs}, these errors appear to stem from the distribution of the load on the sole. Specifically, an excessive force is generated by the module of the fascia and the heel, consequently resulting in less force from the metatarsus.
The error significantly diminishes when the obstacle is in the other positions, remaining below 24.2\% with the lower weight and 31.4\% with the higher weight, except for p6. 
When the obstacle is in p6, the relative force error for the plantar fascia module substantially increases due to its exertion of excessive force in the simulation.
Although the force error is lower in this case compared to when the obstacle is in p1, the relative error is higher since the experimental force is of lower magnitude.

A similar trend in the relative error can be observed across the different obstacle heights. Fig. \ref{fig:e_rel} and Fig. \ref{fig:e_abs} show how the relative error is generally higher for the plantar fascia modules with the lowest obstacle height, i.e., \SI{7}{\milli\meter}, differently from the absolute error. This is because, with a shorter obstacle, the force acting on the plantar fascia module diminishes, significantly amplifying the corresponding relative error. We argue that the lack of sensitivity to low forces does not represent an issue, as the SoftFoot is intended to be used in robotic applications, such as humanoid robot locomotion, where the load on the foot is substantially higher. 

Table \ref{tab:mean_error} summarizes the results obtained in term of mean values of the force relative error for each component, considering all 48 simulations. The mean relative force error is 14.6\% for the heel, 33.2\% for the plantar fascia modules, and 21.7\% for the metatarsus.

\begin{table}
    \centering
    \resizebox{\columnwidth}{!}{%
    \begin{tabular}{|c|c|c|c|}
    \hline
    \textbf{} & \textbf{HEEL} & \textbf{MODULES} & \textbf{METATARSUS} \\
    \hline 
       \textbf{MuJoCo} & 14,6\% & 33,2\% &  21,7\% \\ \hline
        \textbf{MuJoCo Filtered} & 10,5\% & 21,2\% & 13,1\% \\ \hline
        \textbf{Analytical} & 31,0\% & 59,5\% & 7,2\% \\ \hline
        \textbf{Analytical Filtered} & 22,4\% & 54,5\% & 5,2\% \\ \hline
    \end{tabular}%
    }
    
    \caption{Mean relative errors of contact forces estimation with the MuJoCo and the analytical \cite{Mura2020} models, before and after filtering.}
    \label{tab:mean_error}
\end{table}

\begin{figure*}[h]
    \centering
    \begin{minipage}[b]{0.48\textwidth}
        \centering
        \includegraphics[width=\linewidth]{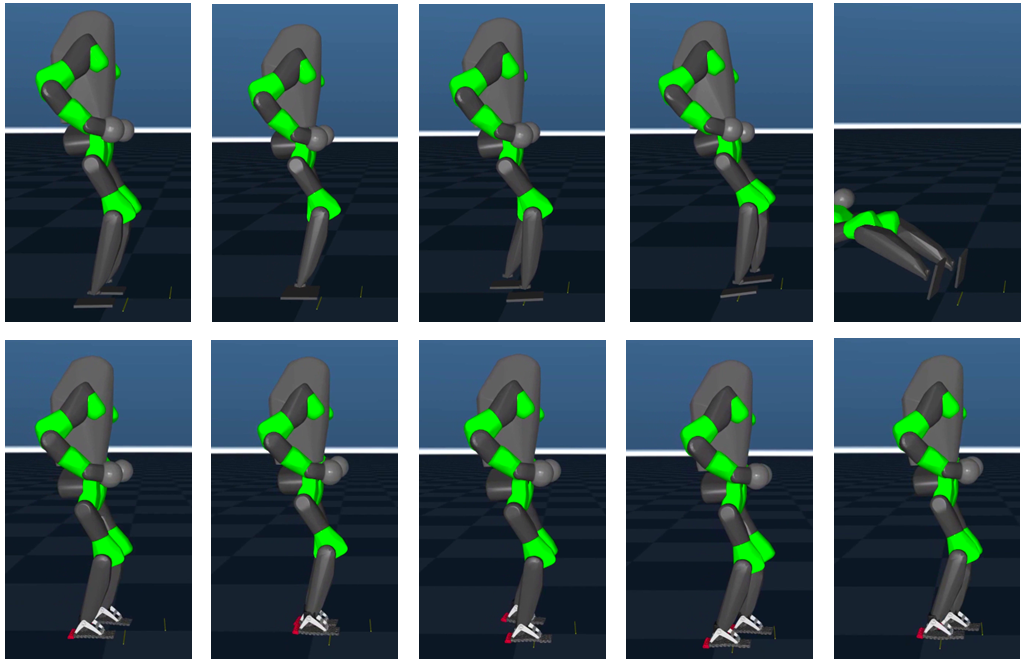}
        \par\vspace{1ex} 
        \text{(a) 5 mm} 
        \label{fig:walk_5}
    \end{minipage}
    \hfill
    \begin{minipage}[b]{0.48\textwidth}
        \centering
        \includegraphics[width=\linewidth]{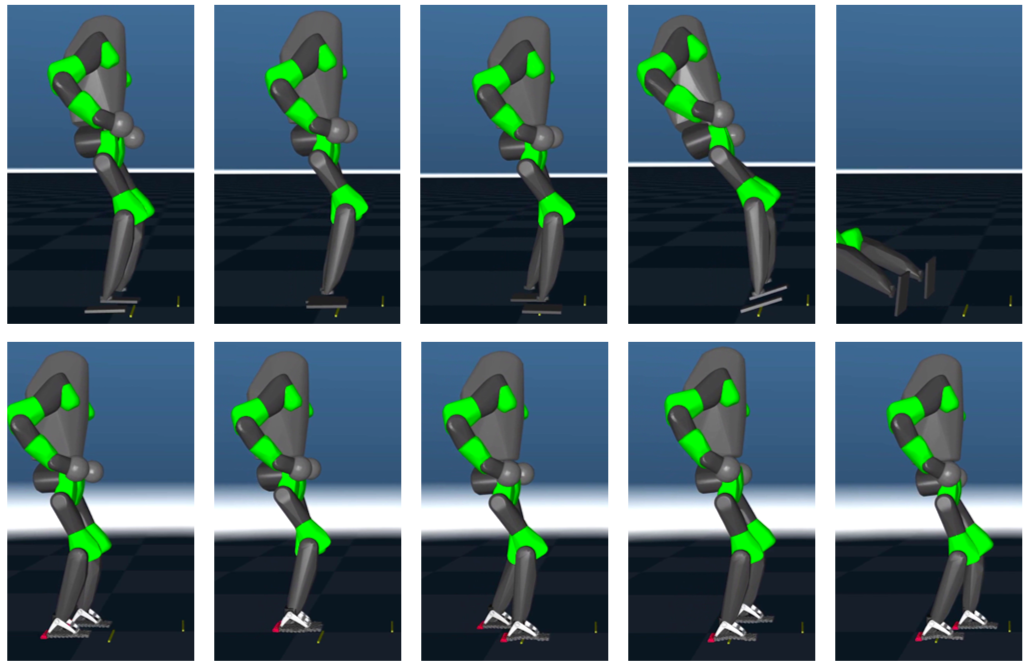}
        \par\vspace{1ex} 
        \text{(b) 10 mm} 
        \label{fig:walk_10}
    \end{minipage}
    \caption{Photo-sequence example of the COMAN+ MuJoCo model (represented in collision geometries) walking on an obstacle, \SI{5}{\milli\meter} in (a), \SI{10}{\milli\meter} in (b),  with its own feet (top) and the SoftFoot (bottom). Notice how the robot is capable of negotiating obstacles when equipped with SoftFoot while it loses balance when using its rigid flat feet.}
    \label{fig:walk}
\end{figure*}

When reconstructing the experimental setup in simulation, we encountered uncertainties regarding the setup itself, specifically the positions of the heel and metatarsus obstacles and the initial rotation of the SoftFoot relative to the slider. We found these three parameters to significantly influence the forces obtained in the simulations. For instance, moving the heel obstacle $\pm$\SI{4}{\milli\meter} caused the mean relative force error to vary by 7\% for the heel, 1.5\% for the plantar fascia, and 13\% for the metatarsus. Similarly, moving the metatarsus obstacle $\pm$\SI{4}{\milli\meter} resulted in changes of 1.5\% for the heel, 10\% for the plantar fascia, and 9\% for the metatarsus. Altering the initial inclination from \SI{1}{\degree} to \SI{5}{\degree} affected the errors by 14\% for the heel, 19\% for the plantar fascia, and 20\% for the metatarsus.
Given these uncertainties, we filtered the relative force error values to remove possible outliers with a mean plus standard deviation filter, reducing the mean relative force error to 10.5\% for the heel, 21.2\% for the plantar fascia, and 13.1\% for the metatarsus (see Table \ref{tab:mean_error}).

Compared to the analytical model developed by Mura et al. \cite{Mura2020}, the MuJoCo model improved the force evaluation accuracy. 
The errors in our simulations are generally lower for the plantar fascia modules, although higher when the obstacle is close to the metatarsus module. For the heel and metatarsus obstacles, the errors are more comparable. Quantitatively, the mean relative error in the analytical model is 31.0\% for the heel, 59.5\% for the plantar fascia modules, and 7.2\% for the metatarsus. Applying the same type of filter we used for our results, these errors are 22.4\%, 54.5\%, and 5.2\%, respectively.
The force estimation errors are reported in Table \ref{tab:mean_error}, where they can be easily compared with those of the MuJoCo model.
Thus, our model generally provides a better estimation of contact forces, particularly for the plantar fascia modules and the heel.

\subsection{Sole Deformation Validation}
As shown in Fig. \ref{fig:e_imu}, the rotational error for each module of the plantar fascia is generally small across all modules, though there are differences in the error range among them. Similar to the forces, the error tends to be higher when the obstacle is positioned closer to the heel for all components. Fig. \ref{fig:e_imu_box} provides a summary of the results across all conditions. The reason for the higher error for m1 and m6 could be the pure rotation assumption between coupled modules and the difficulty in placing the IMUs for those modules in the experiments.
The module rotation error presents some differences also across the different trials, but it has a maximum value of \SI{0.3}{\radian}, and from the box plot in Fig. \ref{fig:e_imu_box} it can be seen that considering the error from the first to the third quartile,  the error range is confined between \SI{-0.1}{\radian} and \SI{0.1}{\radian} for all the modules. Moreover, the full error range is limited within these values for modules m2 to m5.

\subsection{Locomotion Assessment}


   
  
  

Fig. \ref{fig:walk} illustrates COMAN+ walking over obstacles using both its original rigid foot and the SoftFoot.
As the rigid foot acts as a cantilever, and given the height difference in the contact point with the ground and the obstacle, the robot's center of mass is projected outside the support area between the feet. The higher the obstacle, the more the motion of the center of mass increases. The SoftFoot ability to filter the obstacle significantly reduces this phenomenon.

Although the robot was sometimes able to maintain balance with the rigid feet on \SI{5}{\milli\meter} obstacles, it consistently fell on \SI{10}{\milli\meter} obstacles. In contrast, the SoftFoot generally handled both obstacle heights without issue, though the outcome depended on where the sole touched the obstacle. The most robust performance was observed when the contact occurred between the posterior and anterior arches, which is the most flexible part of the sole.

Integrating the SoftFoot digital twin with the COMAN+ model was straightforward, requiring only a substitution in the model’s code for the foot description. This integration was similarly seamless with other bipedal robots, as shown in Fig. \ref{fig:model}. However, since the foot isn’t specifically designed for COMAN+ or any particular robot, minor adjustments may be needed. Additionally, the complexity and numerous components of the SoftFoot slowed down the simulation.

\section{CONCLUSION}



This study presented a validated digital twin of the SoftFoot, developed using the MuJoCo physics engine, and released as an open-source tool for the exploration of new locomotion approaches in simulation. The model accurately captures key characteristics of the SoftFoot, including its ground compliance and the ability to stiffen under load. Additionally, it provides reliable estimates of contact forces during interactions with obstacles, making it a valuable tool for testing new designs and applications.

We demonstrated its integration with other MuJoCo models of legged robots and highlighted its potential benefits for bipedal humanoid robot locomotion, improving stability and adaptability. However, the model complexity slowed simulations and required minor adjustments for compatibility with different robots. The open-source release aims to promote further study and application of the SoftFoot capabilities. Future work will focus on simplifying the model to improve simulation speed while maintaining its accuracy.
This digital twin establishes a foundation for future foot designs, facilitating innovations that could significantly advance the capabilities of humanoid robots in real-world applications. 

\section*{ACKNOWLEDGMENT}

The authors would like to thank Manuel Barbarossa, Eleonora Sguerri and Giovanni Rosato for their help.

\bibliographystyle{ieeetr}
\bibliography{bib}

\end{document}